# MindX: Denoising Mixed Impulse Poisson-Gaussian Noise Using Proximal Algorithms

Mohamed Aly and Wolfgang Heidrich
Visual Computing Center, KAUST

***Abstract*—We present a novel algorithm for blind denoising of images corrupted by mixed impulse, Poisson, and Gaussian noises. The algorithm starts by applying the Anscombe variance-stabilizing transformation to convert the Poisson into white Gaussian noise. Then it applies a combinatorial optimization technique to denoise the mixed impulse Gaussian noise using proximal algorithms. The result is then processed by the inverse Anscombe transform. We compare our algorithm to state of the art methods on standard images, and show its superior performance in various noise conditions.**

## I. Introduction

Mixed Poisson-Gaussian noise arises in many image acquisition settings. For example, CCD and CMOS noise contains both Poisson (signal dependent) and Gaussian (signal independent) components due to the photon-to-electric conversion and the electric-to-digital conversion [1], [2]. It also arises in other photon-limited applications such as astronomy [3], fluorescence microscopy [4], confocal microscopy [5], and low-dosage transmission X-ray CT, PET, and SPECT [6].

In addition, impulse noise is inevitable in digital imaging due to faulty sensor pixels, defected memory devices, or transmission errors [7]. This is especially more important in limited-exposure applications or in restoration of old analog-recorded movies and images, since pixels affected by impulse noise contain no useful information and have to be recovered from the noisy image [8], [9].

In this work, we present MindX, an algorithm for **M**ixed **i**mpulse **n**oise **d**enoising using pro**X**imal algorithms. The algorithm denoises images containing Poisson-Gaussian noise that is also corrupted by impulse noise in *unknown* locations. It consists of three steps: (a) applying a Variance Stabilization Transform (VST) to convert the mixed Poisson-Gaussian noise into white Gaussian noise; (b) performing the denoising in presence of mixed impulse Gaussian noise using proximal algorithms seamlessly combining state of the art denoising algorithms with powerful spatial regularizers; and (c) inverting the VST to restore the original pixel range. We perform extensive experiments using standard images with synthetic noise with varying noise conditions. The results show that our algorithm beats the state of the art in various noise scenarios.

*Contribution :* Our main contribution is a novel algorithm to denoise images corrupted by mixed impulse and Poisson-Gaussian noise, which has not been attempted before up to our knowledge. We show the benefit of the new algorithm and its superior quality over other state of the art approaches using extensive experiments in different conditions.

## II. Related Work

Gaussian noise is ubiquitous in imaging sensors and digital cameras, as it is used to model the noise in various digital to analog conversions and thermal noise. The majority of denoising algorithms assume Gaussian noise either explicitly, or implicitly by minimizing a least square term [10], [11], [12]. Some state of the art methods such as BM3D [11] and Non-Local Means (NLM) [13] exploit the self-similarity between image patches. Others use spatial regularizers, such as Total Variation (TV) regularization, to model natural image statistics [14], [15].

Poisson noise appears in various imaging applications due to the quantum effects of light, especially in photon-limited situations. Many algorithms have been proposed to deal with it. They either model the Poisson likelihood directly [16], [5], [17], [18] or apply a VST, such as the Anscombe Transform [19], [20], to convert the noise into Gaussian noise [21], [19]. Other algorithms handle mixed Poisson-Gaussian noise explicitly [4], [1], [20], [22].

Impulse noise has also been extensively studied, either in isolation [23], [9], [8], [10], [24] or in combination with Gaussian noise [25], [26], [27], [28]. The impulse noise is usually assumed to be sparse, and some form of sparsity-inducing norm is used. Note though that there are some exceptions e.g. [29]. Some methods deal directly with the non-convex $\ell_0$ quasi-norm and solve the problem via alternating minimizations [23], [30], [26], while others use convex relaxations with the $\ell_1$ norm for example [27]. Though the $\ell_0$ quasi-norm is non-convex, these algorithms have generally outperformed other convex-based relaxations.

## III. Noise Model

In this work we focus on mixed impulse and Poisson-Gaussian noise. In particular, we adopt the following general noise model:

$$\mathbf{y}_i = \begin{cases} \mathbf{p}_i + \mathbf{n}_i & \text{for } i \in \Omega \\ \mathbf{i}_i & \text{for } i \notin \Omega \end{cases} \tag{1}$$

where $\mathbf{y} \in \mathbb{R}^N$ is the measured image with $N$ pixels and $\mathbf{y}_i$ is the $i$th pixel, $\mathbf{p} \in \mathbb{R}^N$ is a vector of $N$ independent Poisson-distributed random variables $\mathbf{p}_i \sim \mathcal{P}(\mathbf{x}_i)$ with unknown means $\mathbf{x}_i$ where $\mathbf{x}$ is the underlying clean image, $\mathbf{n} \in \mathbb{R}^N$ is a vector of $N$ independent Gaussian random variables $\mathbf{n}_i \sim \mathcal{N}(0, \sigma^2)$ with variance $\sigma^2$, $\Omega = \{i : \mathbf{i}_i = 0\}$ is the region of the image unaffected by the impulse noise $\mathbf{i}_i$. We assume that the region $\Omega$ is unknown, which is called *blind* inpainting [30].

The impulse noise affects $r \in [0, 100]$ % of the pixels of the image. It can be of two types [23], [30]:

1) **salt-and-pepper** noise that takes on values *white* or *black* pixel values $\mathbf{i}_i \in \{0, p\}$ where $p$ is the peak value, or
2) **random-valued** noise that takes on any value uniformly at random in the interval $[0, p]$ i.e. $\mathbf{i}_i \sim \mathcal{U}(0, p)$.

The former is easier to detect using non-linear filters such as Adaptive Median Filtering [8], while the latter is much harder, though some filters exist such as the Adaptive Center-Weighted Median Filter (ACWMF) [9].

The mean of the Poisson noise is $\mathbb{E}(\mathbf{p}_i) = \mathbf{x}_i$ and its variance is $\mathbb{V}(\mathbf{p}_i) = \mathbf{x}_i$. This means that noise vector $\mathbf{p}$ depends on the unknown signal $\mathbf{x}$, and that's why it is signal-dependent; unlike the Gaussian noise vector $\mathbf{n}$ that is independent of $\mathbf{x}$. It also means that its standard deviation $\text{std}(\mathbf{p}_i) = \sqrt{\mathbf{x}_i}$ i.e. the Poisson noise dominates when the signal value $\mathbf{x}_i$ is small (in photon-limited applications) while the Gaussian noise dominates in other applications, see Fig. 1.

## IV. MindX

MindX consists of these three steps, summarized in Alg. 1:

**Step 1** Converts the mixed Poisson-Gaussian noise into white Gaussian noise using the Generalized Anscombe Transform (GAT) [20] VST.

**Step 2** Denoises the mixed impulse Gaussian noise with a combinatorial algorithm that uses the Chambolle-Pock (CP) Primal-Dual algorithm [14]. In particular, we show how to minimize an objective that seamlessly combines a least square data term, the $\ell_0$ quasi-norm prior for the impulse noise [30], a TV regularizer for the denoised image, and BM3D [11].

**Step 3** Inverts the Anscombe transform [1] to get the original pixel range.

### A. Generalized Anscombe Transform (GAT)

The GAT [20] is a variance stabilizing transformation that takes as input a Poisson-Gaussian random variable $\mathbf{y}_i$ according to Eq. (1) and returns another variable

$$\tilde{\mathbf{y}}_i = \begin{cases} 2\sqrt{\mathbf{y}_i + \frac{3}{8} + \sigma^2} & \text{if } \mathbf{y}_i > -\frac{3}{8} - \sigma^2 \\ 0 & \text{otherwise} \end{cases} \tag{2}$$

such that $\tilde{\mathbf{y}}_i$ is a Gaussian random variable with variance 1. The advantage of GAT is that it precludes the difficulty of dealing with the signal-dependent Poisson noise, and transforms the problem into a mixed impulse noise plus AWGN denoising problem.

### B. Mixed Impulse Gaussian Noise Denoising

We now have an image $\tilde{\mathbf{y}}$ that is corrupted by mixed impulse and Gaussian noise. We propose a formulation similar to [27], [30]. In particular, we solve the following combinatorial problem:

$$\tilde{\mathbf{x}}^\star = \arg\min_{\mathbf{x}} \|\mathbf{x} - \tilde{\mathbf{y}} + \mathbf{z}\|_2^2 + g(\mathbf{K}\mathbf{x}) \text{ s.t. } \|\mathbf{z}\|_0 \le \mu \tag{3}$$

where $g(\mathbf{K}\mathbf{x})$ is the regularization term, $\mathbf{z} \in \mathbb{R}^N$ denotes the value of the impulse noise, $\mu$ is a regularization parameter, and $\|\mathbf{z}\|_0$ is the number of non-zero entries in $\mathbf{z}$. Ideally $\mathbf{z}$ should be zero, but the impulse noise is inevitably present and sparse, and thus we wish to minimize its $\ell_0$ quasi-norm.

Inspired by [31], [30], [15], [23], we combine two state of the art priors in the regularizer: (1) Isotropic Total Variation [14] and (2) BM3D [11]. In particular, we define

$$g(\mathbf{K}\mathbf{x}) = \lambda_1 g_1(\mathbf{K}_1\mathbf{x}) + \lambda_2 g_2(\mathbf{K}_2\mathbf{x}) \tag{4}$$
$$= \lambda_1 \|\nabla\mathbf{x}\|_{2,1} + \lambda_2 h(\mathbf{x}) \tag{5}$$

where $\mathbf{K}_1 = \nabla \in \mathbb{R}^{2N\times N}$ is the matrix that computes the spatial gradient, $\mathbf{K}_2 = \mathbf{I} \in \mathbb{R}^{N\times N}$ is the identity matrix, and $\|\nabla\mathbf{x}\|_{2,1} = \sum_i \|\nabla\mathbf{x}_i\|_2$ is the sum of the magnitudes of the gradient $\nabla\mathbf{x}_i$ at each pixel $i$. The second term in Eq. (4) is the negative log-prior of the image assuming a Gaussian distribution for the noise. The proximal operator for $h(\cdot)$ (see (9)) is equivalent to running any Gaussian denoiser [31], [15], and here we use BM3D. The weights $\lambda_1$ and $\lambda_2$ define the tradeoff between the two priors and the data term. The combined matrix $\mathbf{K} \in \mathbb{R}^{3N\times N}$ is formed by stacking the matrices vertically $\mathbf{K} = [\mathbf{K}_1; \mathbf{K}_2]$ .

Problem (3) is combinatorial and non-convex because of the $\ell_0$ quasi-norm. We explain a model based on the Adaptive Outlier Pursuit [30] to compute a local minimum. It proceeds in two steps:

#### 1. $\mathbf{z}$-*step*

We eliminate $\mathbf{z}$ from (3). Given a value for $\tilde{\mathbf{x}}^{(t)}$ at iteration $t$, then solving (3) for $\mathbf{z}$ is equivalent to solving

$$\mathbf{z}^{(t+1)} = \arg\min_{\mathbf{z}} \|\mathbf{q} - \mathbf{z}\|_2^2 \text{ s.t. } \|\mathbf{z}\|_0 \le \mu \tag{6}$$

which has the closed-form solution [30] for $\mathbf{q} = \tilde{\mathbf{y}} - \tilde{\mathbf{x}}^{(t)}$:

$$\mathbf{z}_i^{(t+1)} = \begin{cases} |\mathbf{q}_i| & \text{if } |\mathbf{q}_i| \ge \mathbf{q}_\mu \\ 0 & \text{otherwise,} \end{cases} \tag{7}$$

where $\mathbf{q}_\mu$ is the entry with the $\mu$th largest magnitude of $\mathbf{q}$ i.e. we find the largest $\mu$ entries of $\text{abs}(\mathbf{q})$. This means that we can deduce the noise-free region $\Omega^{(t)}$ as [26], [30] $\Omega^{(t)} = \{i : \mathbf{z}_i^{(t)} = 0\}$, since when $\mathbf{z}_i^{(t)} \ne 0$ for pixel $i$ the corresponding term in (3) becomes 0.

*2.* $\mathbf{x}$-*step :* Given a fixed value $\mathbf{z}^{(t)}$, and consequently $\Omega^{(t)}$, the problem to solve is equivalent

$$\tilde{\mathbf{x}}^{(t+1)} = \arg\min_{\mathbf{w}} \|\mathbf{w}_{\Omega^{(t)}} - \tilde{\mathbf{y}}_{\Omega^{(t)}}\|_2^2 + g(\mathbf{K}\mathbf{w}) = f(\mathbf{w}) + g(\mathbf{K}\mathbf{w}) \tag{8}$$

where $\mathbf{w}_{\Omega^{(t)}}$ contains only the indices in the "noise free" region $\Omega^{(t)}$ at iteration $t$. This is a non-blind inpainting problem, where we know the values of pixels in $\Omega^{(t)}$ and want to deduce values for damaged pixels outside. It is a convex optimization problem, and we solve it using the CP algorithm[14]. For this, we need the proximal operator for $f(\cdot)$ defined as

$$\text{prox}_{\tau f}(\mathbf{t}) = \arg\min_{\mathbf{w}} f(\mathbf{w}) + \frac{1}{2\tau}\|\mathbf{w} - \mathbf{t}\|_2^2. \tag{9}$$

We also need the proximal operator for $g^*(\cdot)$ the convex conjugate [32] of $g(\cdot)$, which we can find using Moreau's decomposition [32] as

$$\text{prox}_{\rho g^*}(\mathbf{t}) = \mathbf{t} - \rho\, \text{prox}_{\rho^{-1} g}(\rho^{-1}\mathbf{t}).$$

**Algorithm 1** MindX Denoising

**Require:** noisy image $\mathbf{y} \in \mathbb{R}^N$, noise variance $\sigma^2 \in \mathbb{R}$, parameters $\lambda_1, \lambda_2, \mu, \rho, \tau, \theta \in \mathbb{R}$.

▷ ▷ Step 1
$\tilde{\mathbf{y}} = \text{GAT}(\mathbf{y})$ acc. (2).
▷ ▷ Step 2
Initialize: $\mathbf{z}^{(0)} = |\text{AMF}(\mathbf{y}) - \mathbf{y}|$ or $\mathbf{z}^{(0)} = |\text{ACWMF}(\mathbf{y}) - \mathbf{y}|$.
**for** $t = 0 \dots T$ **do**
    $\Omega^{(t)} = \{i : \mathbf{z}^{(t)} = 0\}$.
    ▷ ▷ $\mathbf{x}$-step.
    Initialize $\mathbf{w}^{(0)} = \tilde{\mathbf{x}}^{(t)}$, $\mathbf{u}^{(0)} = \mathbf{K}\mathbf{w}^{(0)}$ and $\mathbf{s}^{(0)} = \mathbf{0}_N$.
    **for** $k = 0 \dots K$ **do**
        $\mathbf{u}^{(k+1)} = \text{prox}_{\rho g^*}\left(\mathbf{u}^{(k)} + \rho\mathbf{K}\mathbf{s}^{(k)}\right)$ acc. (11) and (12).
        $\mathbf{w}^{(k+1)} = \text{prox}_{\tau f}\left(\mathbf{w}^{(k)} - \tau\mathbf{K}^T\mathbf{u}^{(k+1)}\right)$ acc. (14).
        $\mathbf{s}^{(k+1)} = \mathbf{s}^{(k)} + \theta(\mathbf{w}^{(k+1)} - \mathbf{w}^{(k)})$.
    **end for**
    $\tilde{\mathbf{x}}^{(t+1)} = \mathbf{w}^{(K)}$.
    ▷ ▷ $\mathbf{z}$-step.
    $\mathbf{z}^{(t+1)} = \arg\min_{\mathbf{z}} \|\tilde{\mathbf{y}} - \tilde{\mathbf{x}}^{(t+1)} - \mathbf{z}\|_2^2 \text{ s. t. } \|\mathbf{z}\|_0 \leq \mu$ acc. (7).
**end for**
▷ ▷ Step 3
$\mathbf{x}^\star = \text{IGAT}(\tilde{\mathbf{x}}^T)$ according to (15).
**return** clean image estimate $\mathbf{x}^\star$.

We then iterate the following:

1) Solve for $\mathbf{u}^{(k+1)}$ as

$$\mathbf{u}^{(k+1)} = \text{prox}_{\rho g^*}\left(\mathbf{u}^{(k)} + \rho\mathbf{K}\mathbf{s}^{(k)}\right), \quad (10)$$

where $g^*(\mathbf{t})$ is the convex conjugate of $g(\mathbf{t})$ [32]. The proximal operator for $\mathbf{t} = [\mathbf{t}^1 \in \mathbb{R}^{2N}; \mathbf{t}^2 \in \mathbb{R}^N]$ is

$$\text{prox}_{\rho g}(\mathbf{t}) = [\text{prox}_{\rho\lambda_1 g_1}(\mathbf{t}^1);\ \text{prox}_{\rho\lambda_2 g_2}(\mathbf{t}^2)]$$

where

$$\text{prox}_{\rho\lambda_1 g_1}(\mathbf{t}_i^1) = \mathbf{t}_i^1 - \frac{\lambda_1\rho\mathbf{t}_i^1}{\max(\lambda_1\rho, \|\mathbf{t}_i^1\|_2)} \quad (11)$$

$$\text{prox}_{\rho\lambda_2 g_2}(\mathbf{t}^2) = \text{BM3D}(\mathbf{t}^2) \quad (12)$$

and $\mathbf{t}_i^1 \in \mathbb{R}^2$ is the $i$th component of $\mathbf{t}^1$ corresponding to the 2D gradient at pixel $i$. For $g_2(\cdot)$ we just run the BM3D denoising algorithm.

2) Solve for $\mathbf{w}^{(k+1)}$ as

$$\mathbf{w}^{(k+1)} = \text{prox}_{\tau f}\left(\mathbf{w}^{(k)} - \tau\mathbf{K}^T\mathbf{u}^{(k+1)}\right) \quad (13)$$

where the proximal operator for $\mathbf{t} \in \mathbb{R}^N$ is [32]

$$\text{prox}_{\tau f}(\mathbf{t}_i) = \begin{cases} \frac{2\tau}{2\tau+1}\tilde{\mathbf{y}}_i + \frac{1}{2\tau+1}\mathbf{t}_i & \text{for } i \in \Omega^{(t)} \\ \mathbf{t}_i & \text{otherwise.} \end{cases} \quad (14)$$

3) Solve for $\mathbf{s}^{(k+1)}$:

$$\mathbf{s}^{(k+1)} = \mathbf{s}^{(k)} + \theta(\mathbf{w}^{(k+1)} - \mathbf{w}^{(k)}).$$

### C. Inverse GAT (IGAT)

Given the solution $\tilde{\mathbf{x}}$ of the last step, we need to apply the inverse GAT to get the estimate $\mathbf{x}^\star$ in the original pixel range of the input image $\mathbf{y}$. There are different variants, depending on their properties, and here we adopt the *exact unbiased* [1] version that was shown to be better than the straightforward algebraic inverse GAT. In particular, it is defined as

$$\mathbf{x}_i^\star = \int_{-\infty}^{\infty} 2\sqrt{\mathbf{y}_i + \frac{3}{8} + \sigma^2} \sum_{k=0}^{\infty} \left( \frac{\tilde{\mathbf{x}}_i^k e^{-\tilde{\mathbf{x}}_i}}{k!\sqrt{2\pi\sigma^2}} e^{-\frac{(\mathbf{y}_i - k)^2}{2\sigma^2}} \right) d\mathbf{y}_i, \quad (15)$$

and is computed using interpolation from a precomputed look up table; see [1].

*Convergence:* Though Problem (3) is a non-convex combinatorial optimization problem, we have the following result. See Fig. 1 for an illustration.

**Theorem 1.** *Algorithm 1 converges to a local minimum.*

*Proof:* First note that step 2 of Algorithm 1 converges to a local minimum of (3). This follows from [30, Theorem 4.4] and the facts that (a) the $\mathbf{x}$-step to solve for the unique minimizer of (8) converges according to [14, Theorem 1]; and (b) the $\mathbf{z}$-step gives a unique minimum for (6). The main result follows since steps 1 and 3 are non-iterative. ∎

## V. Experiments and Results

### A. Settings

We compare MindX against three state of the art methods, and two variants of MindX:

1) L0TV Proximal ADMM (**L0TV**) [1] [23]: which minimizes an $\ell_0$ quasi-norm data term with a TV regularizer term. It uses a complementary formulation for the $\ell_0$ term. It works best when the impulse noise locations are known in advance i.e. *non-blind inpainting*. They deal only with impulse noise, but not Gaussian or Poisson noises. They show superior performance in this setting against other state of the art methods e.g. [26], [30].

[1] Code at http://yuanganzhao.weebly.com

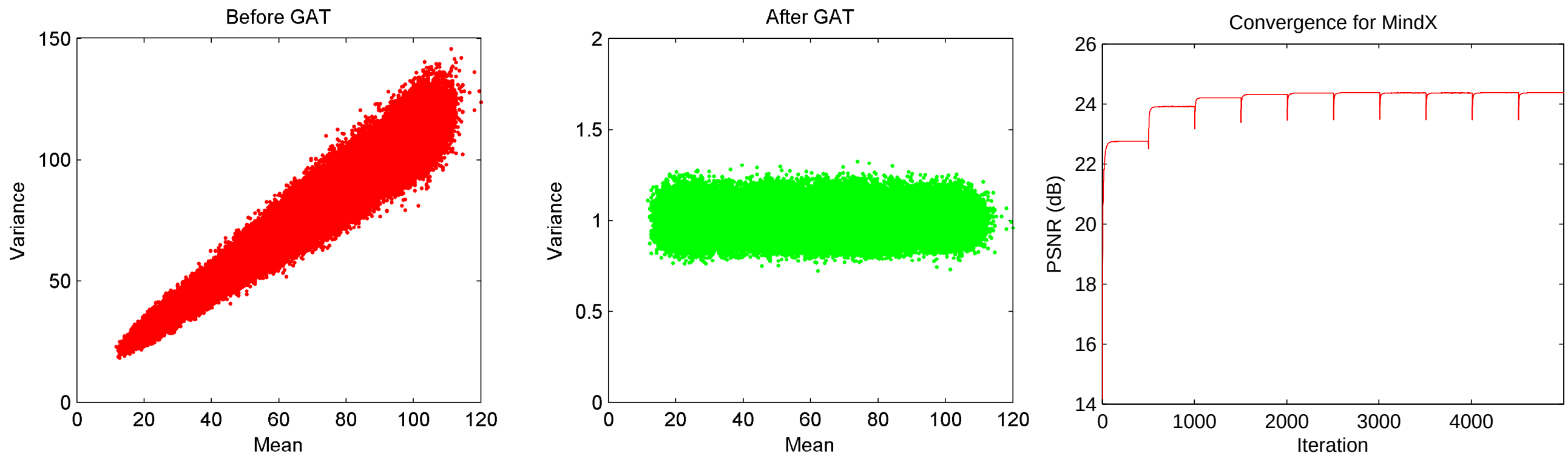

Figure 1. (*left* & *center*) Effect of GAT on stabilizing the variance. Figures show the variance vs. pixel values before (*left*) and after (*right*) GAT. (*right*) PSNR vs. iteration for MindX with random-value noise (10 outer iterations and 100 inner iterations). Note that the algorithm stabilizes after about 6 iterations. The dips at the start of the outer iteration are due to the re-initialization of the CP algorithm. See Sec. IV.

2) Anscombe Transform + BM3D (**GAT-BM3D**) [2] [1]: which performs the GAT on the input noisy image, followed by BM3D Gaussian-noise denoising, and then the Inverse GAT. This method deals only with Poisson-Gaussian noise, but not impulse noise. They show superior performance in this setting against other state of the art methods e.g. [4].
3) Adaptive Outlier Pursuit (**AOP**) [3] [30]: which solves a problem with $\ell_0$ quasi-norm, and follows a two-step alternating minimization. It initializes the first iteration by either AMF or ACWMF to detect the impulse noise. They deal with mixed Gaussian and impulse noise, but not Poisson noise. They show superior results in this setting against other state of the art methods using $\ell_1$ norms e.g. [27] and other methods using the $\ell_0$ term e.g. [26].
4) Anscombe Transform + AOP (**GAT-AOP**) [30]: which combines a GAT preprocessing step (as in GAT-BM3D) followed by AOP to remove mixed Gaussian and impulse noise. This is included as yet another baseline to have a fair comparison with MindX.
5) **MindX-TV**: which uses just TV regularizer in (4) i.e. $\lambda_2 = 0$.
6) **MindX-TV-BM3D**: which uses both TV and BM3D regularizers. We set $\lambda_1 = \lambda_2$ as this gave good results in our experiments.

Since some of the above methods don't deal with impulse noise, we modify the above methods slightly to have a fair comparison. We run AMF/ACWMF (depending on the type of impulse noise) as a preprocessing step before both L0TV and GAT-BM3D. This gives an approximate inpainting mask for L0TV, and removes the impulse noise for GAT-BM3D to get a good baseline.

We run three different experiments with different noise conditions:

1) **Experiment 1**: we scale the input image to have different peak values, as done in [1], and fix the Gaussian noise $\sigma = 0.1p$ where $p$ is the peak value and the amount of pixels affected by impulse noise $r = 50\%$. Different peak values correspond to different amounts of Poisson counts e.g. a low peak value indicates low photon count such as in low-illumination situations.
2) **Experiment 2**: we fix the amount of impulse noise $r = 50\%$ and the peak value $p = 20$ and use different values for the ratio of the standard deviations of the Gaussian/Poisson noises $\sigma/\sqrt{p}$. This corresponds to different amounts of Gaussian noise, from $\sigma = 0$ (pure Poisson noise) to $\sigma = \sqrt{p}$ (equal Poisson and Gaussian noises).
3) **Experiment 3**: we fix the peak value $p = 20$ and the Gaussian noise $\sigma = 0.1p$ and change the amount of impulse noises $r$ from 10% to 70%.

We report our results on four standard $512 \times 512$-pixel images: lena, cameraman, barbara, and pepper. All code was implemented in Matlab, and will be made publicly available. We used code available from the respective authors for the other algorithms. The numbers for AOP, L0TV, and MindX depict the best results obtained using different parameters for all the algorithms, as was done by [23], [30]. Consistent [30], we run 10 outer iterations ($T$ in Alg. 1) for both MindX and AOP for random-value impulse noise, and only 1 for salt-and-pepper noise, as in this case the performance doesn't improve beyond the first iteration. We set $\rho = 500$, $\theta = 1$, and $\tau = 1/(\rho\|\mathbf{K}\|^2)$, and $K = 500$ inner iterations for MindX. We use the default settings for the other algorithms. We report our results using Peak Signal-to-Noise Ratio (PSNR) [19]:

$$\text{PSNR}(\mathbf{x}, \mathbf{x}^\star) = 10 \log_{10} \frac{p^2}{\sum_{i=1}^{N} (\mathbf{x}_i^2 - \mathbf{x}_i^\star)^2 / N} \text{ dB},$$

where $\mathbf{x}^\star$ is the estimate of the ground truth image $\mathbf{x}$ and $p \in [0, 255]$ is the peak pixel value in the experiment.

### B. Results

Fig. 1 shows the effect of the GAT on stabilizing the variance of the Poisson-Gaussian noise and transforming pixel values

[2]Code at http://www.cs.tut.fi/~foi/invansc

[3]Code at https://code.google.com/archive/p/binary-matching-pursuit/downloads

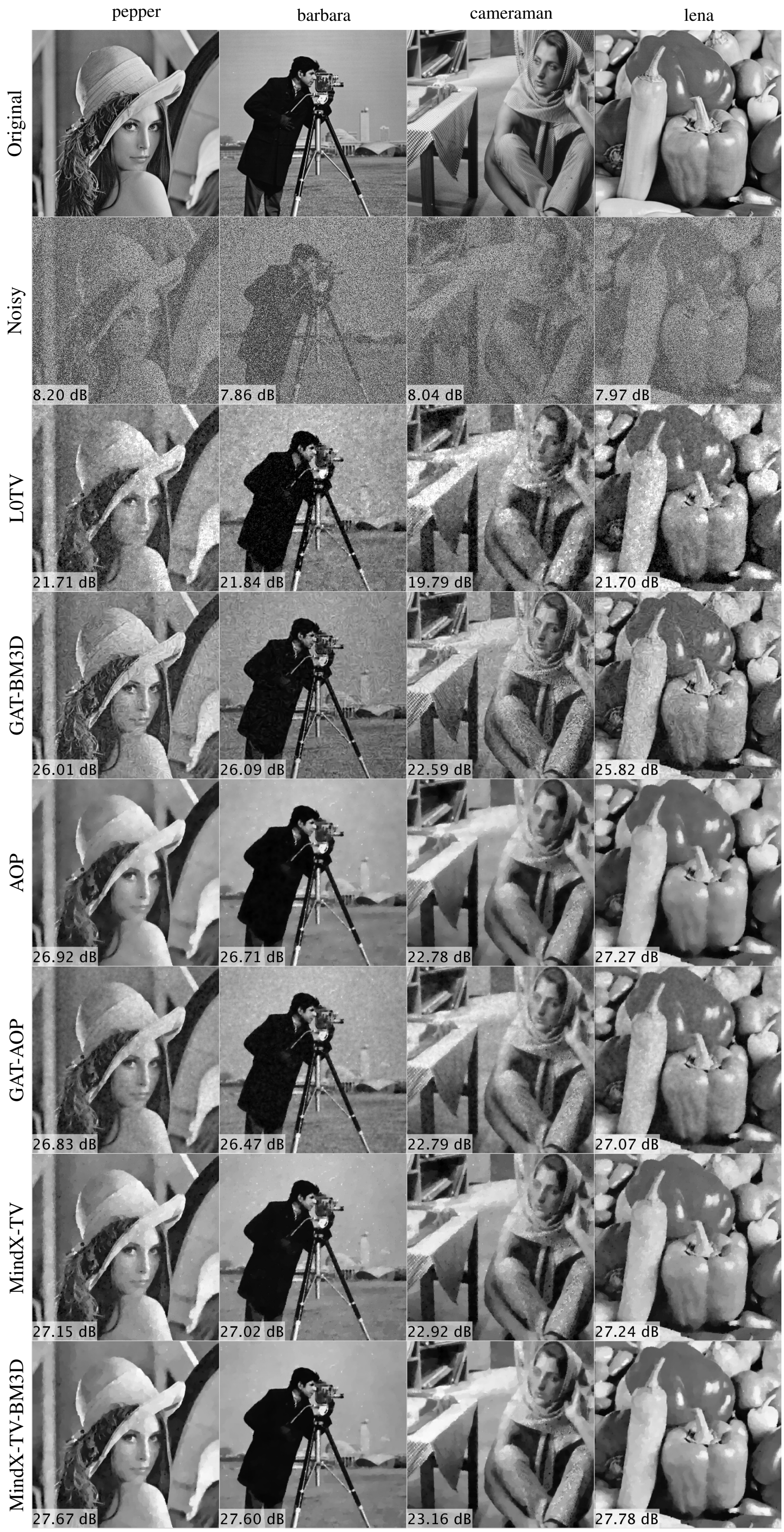

Figure 2. Sample denoising results with $p = 120$, $\sigma = 0.1p$, and $r = 50\%$ salt-and-pepper noise. PSNR is in lower-left corner. The reader can zoom on the images, since they are high resolution.

Table I
EXPERIMENT 1. PSNR VALUES FOR DIFFERENT VALUES OF THE PEAK $p$ FROM 1 TO 120 WITH $\sigma = 0.1p$ AND $r = 50\%$ IMPULSE NOISE. **BOLD GREEN** INDICATES THE BEST VALUES, AND RED INDICATES THE 2ND BEST. SEE SEC. V.

| Noise | Image | lena | | | | | | | | cameraman | | | | | | | |
|---|---|---|---|---|---|---|---|---|---|---|---|---|---|---|---|---|---|
| | Alg. \ $p$ | 1 | 2 | 5 | 10 | 20 | 30 | 60 | 120 | 1 | 2 | 5 | 10 | 20 | 30 | 60 | 120 |
| Salt-and-Pepper Noise | Noisy Input | 3.97 | 5.63 | 6.99 | 7.60 | 7.91 | 8.02 | 8.13 | 8.20 | 4.06 | 5.59 | 6.85 | 7.35 | 7.63 | 7.75 | 7.82 | 7.86 |
| | L0TV | 7.45 | 12.31 | 15.51 | 16.76 | 18.53 | 19.48 | 20.90 | 21.71 | 7.81 | 12.24 | 14.02 | 16.07 | 18.21 | 19.67 | 21.05 | 21.82 |
| | GAT-BM3D | 16.91 | 18.26 | 19.75 | 21.56 | 23.32 | 24.07 | 25.19 | 26.01 | 17.15 | 18.55 | 19.96 | 21.70 | 23.37 | 24.26 | 25.51 | 26.09 |
| | AOP | 16.97 | 18.62 | 20.07 | 20.90 | 24.23 | 25.47 | 26.56 | 26.92 | 16.94 | 18.50 | 19.91 | 20.76 | 23.86 | 25.22 | 26.40 | 26.71 |
| | GAT-AOP | 18.61 | 20.23 | 22.54 | 24.09 | 24.86 | 25.81 | 26.45 | 26.83 | 18.56 | 19.97 | 22.04 | 23.63 | 24.49 | 25.52 | 26.25 | 26.47 |
| | MindX-TV | 18.96 | 20.45 | 22.80 | 24.24 | 25.49 | 26.03 | 26.69 | 27.15 | 18.80 | 20.12 | 22.58 | 24.04 | 25.10 | 25.88 | 26.61 | 27.02 |
| | MindX-TV-BM3D | **19.19** | **21.54** | **23.93** | **24.95** | **26.22** | **26.83** | **27.46** | **27.67** | **19.31** | **21.38** | **23.83** | **24.89** | **26.25** | **26.95** | **27.55** | **27.60** |
| | Image | barbara | | | | | | | | pepper | | | | | | | |
| | Alg. \ $p$ | 1 | 2 | 5 | 10 | 20 | 30 | 60 | 120 | 1 | 2 | 5 | 10 | 20 | 30 | 60 | 120 |
| | Noisy Input | 4.03 | 5.64 | 6.93 | 7.47 | 7.78 | 7.88 | 7.98 | 8.04 | 3.78 | 5.42 | 6.79 | 7.37 | 7.69 | 7.80 | 7.90 | 7.97 |
| | L0TV | 7.71 | 10.80 | 13.96 | 15.34 | 17.19 | 17.97 | 19.11 | 19.80 | 7.86 | 12.27 | 14.60 | 16.00 | 18.06 | 19.23 | 20.71 | 21.69 |
| | GAT-BM3D | 16.49 | 17.45 | 18.67 | 19.82 | 20.88 | 21.46 | 22.13 | 22.59 | 16.55 | 18.06 | 19.93 | 21.60 | 23.31 | 24.13 | 25.15 | 25.82 |
| | AOP | 16.05 | 17.12 | 18.13 | 19.54 | 21.51 | 22.12 | 22.62 | 22.78 | 15.92 | 17.36 | 18.76 | 21.11 | 24.25 | 25.58 | 26.87 | 27.27 |
| | GAT-AOP | 17.69 | 18.95 | 20.58 | 21.34 | 21.80 | 22.20 | 22.54 | 22.80 | 17.88 | 19.89 | 22.59 | 24.26 | 25.07 | 26.02 | 26.86 | 27.07 |
| | MindX-TV | 17.69 | 18.80 | 20.49 | 21.33 | 21.91 | 22.27 | 22.65 | 22.95 | 17.99 | 19.55 | 22.28 | 24.06 | 25.39 | 26.02 | 26.87 | 27.24 |
| | MindX-TV-BM3D | **18.03** | **19.64** | **21.25** | **21.74** | **22.33** | **22.68** | **23.04** | **23.23** | **18.42** | **20.91** | **23.80** | **25.01** | **26.23** | **26.85** | **27.60** | **27.78** |
| Random-Value Noise | Image | lena | | | | | | | | cameraman | | | | | | | |
| | Alg. \ $p$ | 1 | 2 | 5 | 10 | 20 | 30 | 60 | 120 | 1 | 2 | 5 | 10 | 20 | 30 | 60 | 120 |
| | Noisy Input | 4.98 | 7.21 | 9.34 | 10.42 | 11.07 | 11.29 | 11.53 | 11.69 | 5.09 | 7.14 | 9.07 | 9.98 | 10.51 | 10.72 | 10.91 | 11.01 |
| | L0TV | 11.02 | 15.18 | 17.09 | 18.06 | 19.42 | 20.05 | 21.10 | 21.57 | 11.13 | 16.18 | 17.20 | 18.42 | 19.33 | 20.07 | 20.25 | 20.90 |
| | GAT-BM3D | 17.60 | 17.65 | 18.87 | 19.80 | 20.51 | 20.79 | 21.25 | 21.61 | 16.87 | 17.34 | 18.20 | 18.81 | 19.27 | 19.55 | 19.69 | 19.75 |
| | AOP | 13.64 | 16.12 | 18.50 | 19.71 | 21.19 | 22.14 | 23.41 | 23.92 | 14.61 | 17.06 | 18.70 | 20.04 | 20.97 | 21.69 | 22.55 | 22.94 |
| | GAT-AOP | 14.24 | 15.68 | 18.75 | 20.48 | 21.90 | 22.46 | 23.25 | 23.50 | 14.52 | 17.75 | 18.86 | 20.28 | 21.20 | 21.89 | 22.29 | 22.35 |
| | MindX-TV | **17.89** | **18.47** | 18.91 | 20.56 | 22.14 | 22.67 | 23.58 | 23.97 | **17.17** | **17.98** | 19.40 | 20.54 | 21.64 | 22.19 | 22.70 | 23.03 |
| | MindX-TV-BM3D | 14.03 | 15.41 | **19.06** | **21.02** | **22.72** | **23.25** | **24.17** | **24.37** | 13.55 | 17.71 | **19.52** | **21.08** | **22.40** | **22.91** | **23.54** | **23.58** |
| | Image | barbara | | | | | | | | pepper | | | | | | | |
| | Alg. \ $p$ | 1 | 2 | 5 | 10 | 20 | 30 | 60 | 120 | 1 | 2 | 5 | 10 | 20 | 30 | 60 | 120 |
| | Noisy Input | 5.07 | 7.24 | 9.23 | 10.19 | 10.79 | 10.98 | 11.22 | 11.34 | 4.75 | 6.91 | 8.98 | 10.01 | 10.60 | 10.82 | 11.05 | 11.18 |
| | L0TV | 11.25 | 14.88 | 16.18 | 17.20 | 18.33 | 18.62 | 19.37 | 19.69 | 10.88 | 14.16 | 16.12 | 17.37 | 18.81 | 19.57 | 20.38 | 20.98 |
| | GAT-BM3D | **16.71** | 16.35 | 17.41 | 18.06 | 18.68 | 18.91 | 19.20 | 19.38 | **17.01** | 16.98 | 17.88 | 18.65 | 19.44 | 19.65 | 20.11 | 20.35 |
| | AOP | 13.27 | 15.63 | 17.46 | 18.41 | 19.50 | 20.06 | 20.74 | 21.06 | 13.23 | 15.00 | 17.64 | 18.95 | 20.46 | 21.33 | 22.68 | 23.36 |
| | GAT-AOP | 13.58 | 15.05 | 17.58 | 18.95 | 19.87 | 20.29 | 20.75 | 20.90 | 12.99 | 15.53 | 18.33 | 19.88 | 21.31 | 21.85 | 22.60 | 22.93 |
| | MindX-TV | 16.37 | **16.84** | 17.64 | 18.98 | 19.96 | 20.35 | 20.82 | 21.06 | 16.93 | **17.40** | 18.39 | 19.92 | 21.46 | 21.94 | 22.82 | 23.30 |
| | MindX-TV-BM3D | 13.63 | 14.54 | **17.76** | **19.32** | **20.38** | **20.69** | **21.16** | **21.28** | 12.38 | 15.08 | **18.67** | **20.54** | **22.28** | **22.68** | **23.52** | **23.80** |

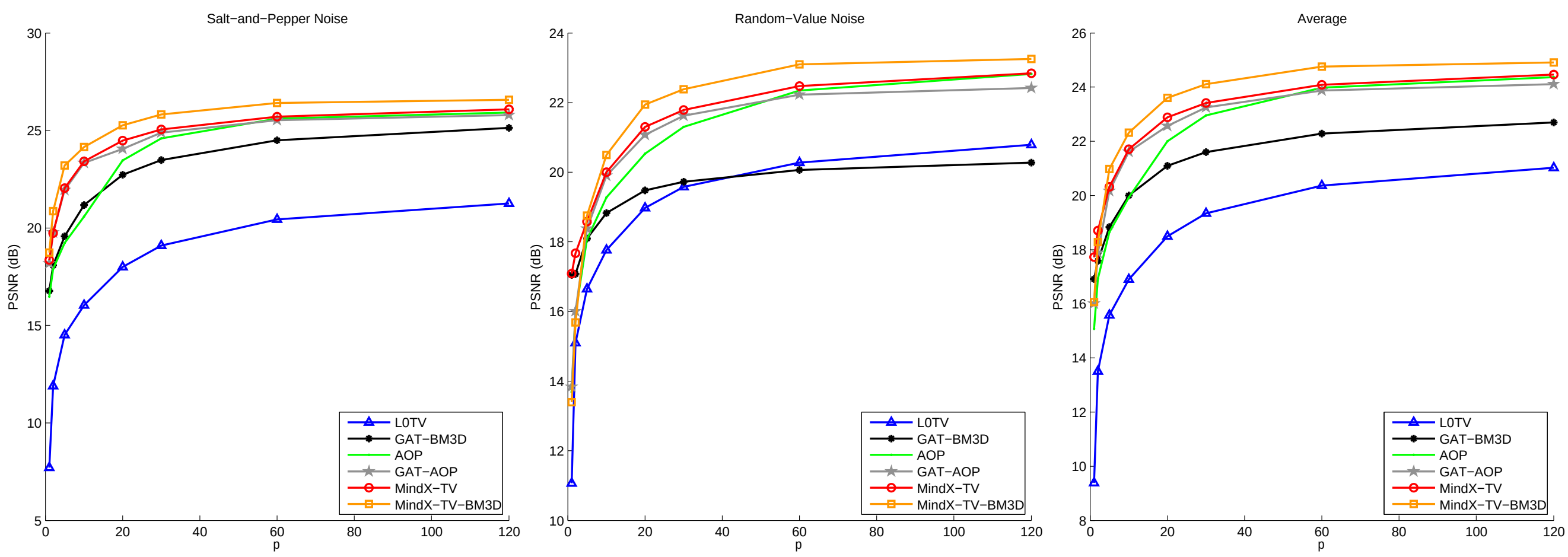

Figure 3. Average PSNR results over the images for Experiment 1. See Sec. V and Table I.

Table II

EXPERIMENT 2. PSNR VALUES FOR DIFFERENT VALUES OF THE RATIO $\sigma/\sqrt{p}$ FROM 0 TO 5 WITH PEAK $p = 20$ AND $r = 50\%$ IMPULSE NOISE. **BOLD GREEN** INDICATES THE BEST VALUES, AND RED INDICATES THE 2ND BEST. SEE SEC. V.

| Salt-and-Pepper Noise | Image | lena | | | | | | | cameraman | | | | | | |
|---|---|---|---|---|---|---|---|---|---|---|---|---|---|---|---|
| | Alg. \ $\sigma/\sqrt{p}$ | 0 | 0.05 | 0.1 | 0.5 | 1 | 2 | 5 | 0 | 0.05 | 0.1 | 0.5 | 1 | 2 | 5 |
| | Noisy Input | 8.04 | 8.04 | 8.04 | 7.88 | 7.40 | 5.91 | 1.08 | 7.76 | 7.76 | 7.75 | 7.60 | 7.16 | 5.73 | 1.03 |
| | L0TV | 19.73 | 19.70 | 19.64 | 18.25 | 16.34 | 14.52 | 15.49 | 19.66 | 19.53 | 19.48 | 17.92 | 15.56 | 13.02 | 14.87 |
| | GAT-BM3D | 24.51 | 24.39 | 24.36 | 23.10 | 20.92 | 18.33 | 16.60 | 24.35 | 24.42 | 24.39 | 23.19 | 20.86 | 18.00 | 15.47 |
| | AOP | 25.61 | 25.46 | 25.43 | 23.86 | 20.24 | 19.06 | 17.38 | 24.16 | 25.21 | 25.17 | 23.54 | 19.74 | 18.66 | 16.30 |
| | GAT-AOP | 25.66 | 25.59 | 25.58 | 24.71 | 23.30 | 20.56 | 15.33 | 24.83 | 25.50 | 25.47 | 24.33 | 22.41 | 19.60 | 14.46 |
| | MindX-TV | 25.94 | 26.09 | 26.07 | 25.34 | 23.92 | 21.52 | **17.61** | 25.75 | 26.05 | 26.01 | 24.92 | 23.11 | 20.83 | 16.31 |
| | MindX-TV-BM3D | **27.05** | **27.14** | **27.12** | **26.08** | **24.67** | **22.00** | 17.60 | **27.28** | **27.34** | **27.32** | **26.08** | **23.99** | **21.26** | **16.36** |

| Salt-and-Pepper Noise | Image | barbara | | | | | | | pepper | | | | | | |
|---|---|---|---|---|---|---|---|---|---|---|---|---|---|---|---|
| | Alg. \ $\sigma/\sqrt{p}$ | 0 | 0.05 | 0.1 | 0.5 | 1 | 2 | 5 | 0 | 0.05 | 0.1 | 0.5 | 1 | 2 | 5 |
| | Noisy Input | 7.91 | 7.91 | 7.90 | 7.75 | 7.29 | 5.83 | 1.05 | 7.82 | 7.82 | 7.82 | 7.66 | 7.21 | 5.77 | 1.03 |
| | L0TV | 18.44 | 18.30 | 18.26 | 16.99 | 15.20 | 13.46 | 14.20 | 19.58 | 19.33 | 19.28 | 17.80 | 15.07 | 13.69 | 14.12 |
| | GAT-BM3D | 21.66 | 21.61 | 21.60 | 20.73 | 19.26 | 17.23 | 15.58 | 24.38 | 24.29 | 24.28 | 23.11 | 20.95 | 18.16 | 15.84 |
| | AOP | 22.17 | 22.19 | 22.17 | 21.32 | 18.31 | 17.41 | 16.15 | 25.27 | 25.70 | 25.66 | 23.89 | 19.47 | 17.92 | 16.26 |
| | GAT-AOP | 22.38 | 22.31 | 22.30 | 21.71 | 20.82 | 19.07 | 14.59 | 25.91 | 25.97 | 25.96 | 24.92 | 23.22 | 20.08 | 14.83 |
| | MindX-TV | 22.34 | 22.36 | 22.35 | 21.83 | 20.99 | 19.44 | **16.34** | 26.12 | 26.32 | 26.29 | 25.21 | 23.42 | 20.73 | **16.52** |
| | MindX-TV-BM3D | **22.89** | **22.89** | **22.87** | **22.25** | **21.44** | **19.75** | 16.32 | **27.32** | **27.24** | **27.22** | **26.07** | **24.30** | **21.33** | **16.52** |

| Random-Value Noise | Image | lena | | | | | | | cameraman | | | | | | |
|---|---|---|---|---|---|---|---|---|---|---|---|---|---|---|---|
| | Alg. \ $\sigma/\sqrt{p}$ | 0 | 0.05 | 0.1 | 0.5 | 1 | 2 | 5 | 0 | 0.05 | 0.1 | 0.5 | 1 | 2 | 5 |
| | Noisy Input | 11.36 | 11.35 | 11.34 | 11.00 | 10.08 | 7.62 | 1.57 | 10.76 | 10.75 | 10.74 | 10.45 | 9.63 | 7.36 | 1.51 |
| | L0TV | 20.37 | 20.24 | 20.15 | 19.16 | 17.81 | 16.26 | 15.21 | 20.26 | 20.23 | 20.16 | 19.23 | 17.41 | 15.31 | 13.50 |
| | GAT-BM3D | 21.08 | 21.08 | 21.05 | 20.38 | 19.35 | 17.73 | 16.01 | 19.82 | 19.83 | 19.81 | 19.14 | 17.87 | 16.04 | 14.19 |
| | AOP | 20.21 | 22.36 | 22.34 | 20.98 | 19.89 | 17.27 | 14.94 | 20.38 | 22.26 | 22.25 | 20.73 | 19.21 | 15.89 | 14.34 |
| | GAT-AOP | 20.34 | 22.97 | 22.88 | 21.67 | 19.83 | 17.12 | 14.62 | 20.21 | 22.48 | 22.42 | 21.02 | 18.60 | 15.63 | 13.76 |
| | MindX-TV | 23.10 | 23.06 | 23.01 | 21.96 | 20.06 | **18.59** | **17.78** | 22.83 | 22.74 | 22.70 | 21.41 | 19.12 | **16.94** | **16.33** |
| | MindX-TV-BM3D | **23.95** | **23.92** | **23.81** | **22.55** | **20.47** | 17.59 | 15.85 | **23.69** | **23.62** | **23.61** | **22.14** | **19.47** | 16.19 | 14.85 |

| Random-Value Noise | Image | barbara | | | | | | | pepper | | | | | | |
|---|---|---|---|---|---|---|---|---|---|---|---|---|---|---|---|
| | Alg. \ $\sigma/\sqrt{p}$ | 0 | 0.05 | 0.1 | 0.5 | 1 | 2 | 5 | 0 | 0.05 | 0.1 | 0.5 | 1 | 2 | 5 |
| | Noisy Input | 11.06 | 11.05 | 11.04 | 10.73 | 9.86 | 7.49 | 1.54 | 10.86 | 10.85 | 10.84 | 10.54 | 9.70 | 7.40 | 1.51 |
| | L0TV | 18.75 | 18.76 | 18.70 | 18.14 | 16.73 | 15.15 | 13.83 | 19.81 | 19.67 | 19.55 | 18.75 | 16.86 | 15.01 | 13.37 |
| | GAT-BM3D | 19.10 | 19.10 | 19.08 | 18.59 | 17.74 | 16.35 | 14.75 | 19.92 | 19.93 | 19.91 | 19.30 | 18.18 | 16.58 | 14.66 |
| | AOP | 18.83 | 20.16 | 20.14 | 19.32 | 18.34 | 16.05 | 13.93 | 19.64 | 21.68 | 21.63 | 20.19 | 18.81 | 16.14 | 13.45 |
| | GAT-AOP | 18.93 | 20.61 | 20.55 | 19.73 | 18.29 | 15.89 | 14.01 | 19.83 | 22.34 | 22.28 | 21.12 | 18.91 | 15.86 | 14.05 |
| | MindX-TV | 20.63 | 20.62 | 20.57 | 19.83 | 18.35 | **17.01** | **16.43** | 22.48 | 22.48 | 22.46 | 21.23 | 19.13 | **17.31** | **16.58** |
| | MindX-TV-BM3D | **21.04** | **21.04** | **21.03** | **20.22** | **18.65** | 16.18 | 14.71 | **23.55** | **23.56** | **23.51** | **22.05** | **19.62** | 16.48 | 14.42 |

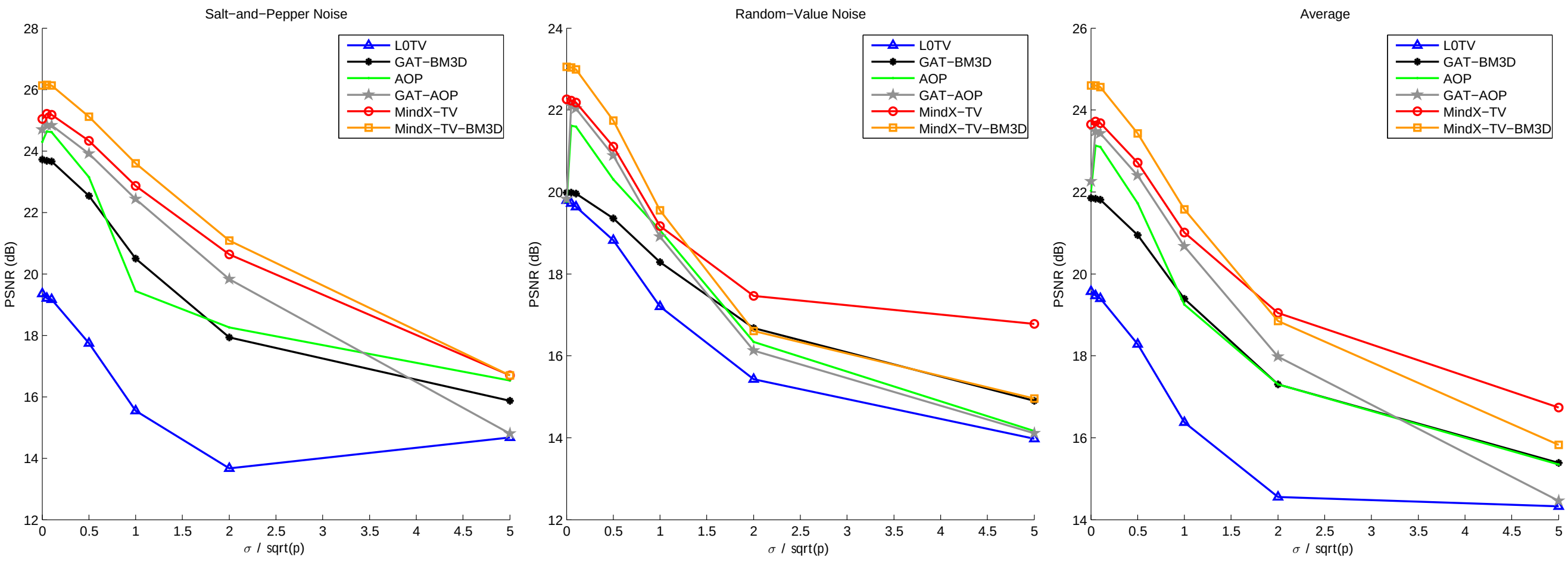

Figure 4. Average PSNR results over the images for Experiment 2. See Sec. V and Table II.

Table III

EXPERIMENT 3. PSNR VALUES FOR DIFFERENT AMOUNTS OF IMPULSE NOISE $r$ FROM 10% TO 90% WITH PEAK $p = 20$ AND $\sigma = 0.1p$. **BOLD GREEN** INDICATES THE BEST VALUES, AND RED INDICATES THE 2ND BEST. SEE SEC. V.

| Noise | Image | **lena** | | | | | **cameraman** | | | | |
|---|---|---|---|---|---|---|---|---|---|---|---|
| | Alg. \ $r$ | 10 | 30 | 50 | 70 | 90 | 10 | 30 | 50 | 70 | 90 |
| Salt-and-Pepper Noise | Noisy Input | 12.20 | 9.54 | 7.91 | 6.73 | 5.79 | 12.17 | 9.33 | 7.63 | 6.44 | 5.48 |
| | L0TV | 20.62 | 19.52 | 18.53 | 17.07 | 14.62 | 20.07 | 19.38 | 18.21 | 16.75 | 14.26 |
| | GAT-BM3D | 27.25 | 25.56 | 23.32 | 20.53 | 17.13 | 27.76 | 25.95 | 23.37 | 20.49 | 17.03 |
| | AOP | 23.90 | 24.05 | 24.23 | 22.37 | 18.01 | 24.13 | 24.40 | 23.86 | 21.73 | 17.90 |
| | GAT-AOP | 26.20 | 25.59 | 24.86 | 23.14 | 17.52 | 26.12 | 25.29 | 24.49 | 22.23 | 16.79 |
| | MindX-TV | 26.62 | 26.03 | 25.49 | **23.88** | 19.97 | 26.90 | 26.20 | 25.10 | **23.18** | 19.76 |
| | MindX-TV-BM3D | **27.58** | **26.91** | **26.22** | 23.53 | **20.42** | **28.33** | **27.55** | **26.25** | 22.96 | **20.02** |
| | Image | **barbara** | | | | | **pepper** | | | | |
| | Alg. \ $r$ | 10 | 30 | 50 | 70 | 90 | 10 | 30 | 50 | 70 | 90 |
| | Noisy Input | 12.20 | 9.44 | 7.78 | 6.59 | 5.64 | 12.03 | 9.32 | 7.69 | 6.50 | 5.56 |
| | L0TV | 18.97 | 18.02 | 17.19 | 16.16 | 13.74 | 20.39 | 19.33 | 18.06 | 16.96 | 14.56 |
| | GAT-BM3D | **23.74** | 22.57 | 20.88 | 18.84 | 16.09 | 27.09 | 25.48 | 23.31 | 20.66 | 17.32 |
| | AOP | 21.83 | 21.72 | 21.51 | 20.11 | 16.46 | 23.94 | 24.34 | 24.25 | 22.26 | 17.05 |
| | GAT-AOP | 22.30 | 22.06 | 21.80 | 20.51 | 15.98 | 26.63 | 25.85 | 25.07 | 22.85 | 17.21 |
| | MindX-TV | 22.59 | 22.25 | 21.91 | **20.94** | **18.42** | 26.84 | 26.12 | 25.39 | **23.55** | 19.91 |
| | MindX-TV-BM3D | 23.19 | **22.79** | **22.33** | 20.79 | 18.34 | **27.79** | **27.15** | **26.23** | 23.23 | **19.97** |
| Random-Value Noise | Image | **lena** | | | | | **cameraman** | | | | |
| | Alg. \ $r$ | 10 | 30 | 50 | 70 | 90 | 10 | 30 | 50 | 70 | 90 |
| | Noisy Input | 13.59 | 12.15 | 11.07 | 10.22 | 9.48 | 13.54 | 11.77 | 10.51 | 9.54 | 8.74 |
| | L0TV | 20.46 | 20.26 | 19.42 | 17.48 | 15.21 | 20.21 | 20.15 | 19.33 | 17.01 | **13.24** |
| | GAT-BM3D | 26.46 | 23.90 | 20.51 | 17.34 | 14.65 | 26.78 | 23.74 | 19.27 | 15.41 | 12.65 |
| | AOP | 24.66 | 23.93 | 21.19 | 18.65 | **15.38** | 24.81 | 23.86 | 20.97 | 17.37 | 13.11 |
| | GAT-AOP | 25.67 | 23.84 | 21.90 | 18.69 | 15.33 | 25.57 | 23.43 | 21.20 | 16.99 | 13.05 |
| | MindX-TV | 25.96 | 24.26 | 22.14 | 18.71 | 15.33 | 26.03 | 24.06 | 21.64 | 17.07 | 13.01 |
| | MindX-TV-BM3D | **27.07** | **25.00** | **22.72** | **19.07** | 15.36 | **27.54** | **25.10** | **22.40** | **17.38** | 13.04 |
| | Image | **barbara** | | | | | **pepper** | | | | |
| | Alg. \ $r$ | 10 | 30 | 50 | 70 | 90 | 10 | 30 | 50 | 70 | 90 |
| | Noisy Input | 13.60 | 11.97 | 10.79 | 9.86 | 9.09 | 13.35 | 11.76 | 10.60 | 9.69 | 8.93 |
| | L0TV | 18.96 | 18.86 | 18.33 | 16.52 | 14.06 | 20.50 | 19.89 | 18.81 | 16.52 | 13.47 |
| | GAT-BM3D | 22.58 | 21.16 | 18.68 | 15.85 | 13.46 | 26.33 | 23.32 | 19.44 | 15.88 | 13.19 |
| | AOP | 21.93 | 21.20 | 19.50 | 17.17 | **14.07** | 24.53 | 23.79 | 20.46 | 17.05 | 13.60 |
| | GAT-AOP | 22.27 | 21.18 | 19.87 | 17.18 | 13.96 | 25.96 | 23.69 | 21.31 | 17.15 | 13.60 |
| | MindX-TV | 22.38 | 21.40 | 19.96 | 17.09 | 13.97 | 26.19 | 24.08 | 21.46 | 17.15 | 13.64 |
| | MindX-TV-BM3D | **22.93** | **21.73** | **20.38** | **17.38** | 14.01 | **27.29** | **24.93** | **22.28** | **17.47** | **13.66** |

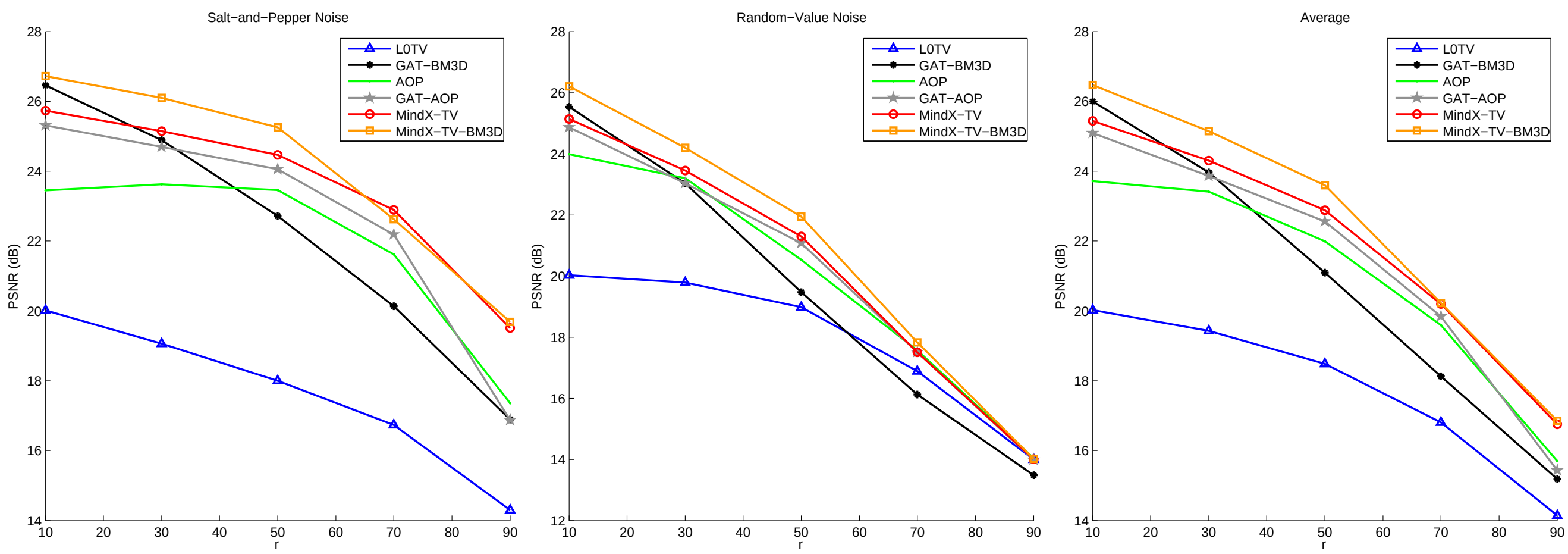

Figure 5. Average PSNR results over the images for Experiment 3. See Sec. V and Table III.

to standard normal variables. It also shows the convergence of MindX with random-value noise. Fig. 2 shows some image results with salt-and-pepper noise. Tables I, II, and III show the results of Experiments 1, 2, and 3, respectively. Figures 3, 4, and 5 show the average PSNR results over the images for Experiments 1, 2, and 3, respectively.

We note that MindX is better than all other algorithms in most noise situations, and is in 1st place in all but 6 settings, where it was in 2nd place in 4 of them. For Experiment 1 (Table I and Fig. 3), GAT-AOP is a close runner-up, and GAT-BM3D is also close especially with lower peak values. For Experiment 2 (Table II and Fig. 4), AOP and GAT-AOP are the close runner-ups with the salt-and-pepper noise. We also note that for extreme values of $\sigma/\sqrt{p}$, MindX-TV is better than MindX-TV-BM3D. We think this is because in these extreme situations the tradeoff between the TV weight $\lambda_1$ and the BM3D self-similarity weight $\lambda_2$ in 4 needs to be better tuned. For Experiment 3 (Table III and Fig. 5), GAT-BM3D is again the runner up with lower impulse noise values and L0TV is competitive with higher impulse noise values. This shows that MindX achieves consistently very well in different noise scenarios, and is generally followed by other algorithms in specific situations.

We also note that adding the BM3D prior to the TV prior generally improves the performance over algorithms that use just the TV prior (L0TV, AOP, GAT-AOP, MindX-TV) or that use the BM3D prior (GAT-BM3D). This shows that it is beneficial to add different priors with different complementary properties.

L0TV is generally worse than AOP, which can be attributed to the fact that it was designed only for impulse noise, as the experiments in [23] show. GAT-BM3D provides a good baseline, and is quite competitive with more sophisticated algorithms such as AOP, especially in low photon-count situations. GAT-AOP is generally better than AOP, especially when the Poisson noise dominates and in low photo-count situations. Random-value impulse noise is harder to remove than salt-and-pepper noise, and PSNR values are lower in that case, although MindX and AOP use more outer iterations in this case (10 vs 1) [30].

## VI. Conclusions

In this work we presented a novel algorithm to solve the problem of denoising images corrupted by both impulse and Poisson-Gaussian noises. The algorithm solves a combinatorial optimization problem involving the $\ell_0$ quasi-norm, in addition to seamlessly combining the TV and BM3D priors. We proved its convergence to a local minimum. We compared it to state of the art denoising methods and showed its superior quality in various noise conditions. We plan to investigate next how to general deconvolution problems, both blind and non-blind. We also plan to integrate it in other applications such as fluorescence microscopy or low-dose X-ray tomography.